\documentclass{article}
\usepackage{spconf,amsmath,amssymb,graphicx,hyperref,booktabs,multirow}

\title{FastEagle: Cascaded Drafting for Accelerating Speculative Decoding}
\name{Haiduo Huang, Jiangcheng Song, Wenzhe Zhao, Pengju Ren}
\address{Institute of Artificial Intelligence and Robotics, Xi'an Jiaotong University}
\begin{document}
%
\maketitle
\begin{abstract}
Speculative decoding accelerates generation by drafting candidates and verifying them in parallel, yet state-of-the-art drafters (e.g., EAGLE) still require $N$ sequential passes to propose $N$ tokens. We present FastEagle, a non-autoregressive cascaded drafter that emits an entire draft in a single forward pass. FastEagle replaces temporal steps with a lightweight layer cascade and trains with layer-wise supervision to mitigate error accumulation. Coupled with a constrained draft tree that preserves lossless verification cost, FastEagle delivers substantial wall-clock speedups over strong autoregressive drafters while maintaining competitive acceptance behavior. Across multiple LLMs (Vicuna-13B, LLaMA-Instruct 3.x, and DeepSeek-R1-Distill-LLaMA) and tasks (MT-Bench, HumanEval, GSM8K, CNN/DM, Alpaca), FastEagle consistently outperforms EAGLE-3 in speedup under both greedy and stochastic decoding, with comparable average acceptance lengths. These results indicate that removing sequential dependencies in drafting is a practical path toward lossless LLM inference acceleration.
\end{abstract}
\section{Introduction}
\label{sec:intro}
Modern Large Language Models (LLMs) demonstrate remarkable capabilities across diverse domains \cite{achiam2023gpt,dubey2024llama,liu2024deepseek,team2024gemini,yang2025qwen3}, yet their inference remains expensive due to the inherently sequential, autoregressive decoding process. Each token is generated conditioned on the previous one, creating a latency bottleneck that hinders real-time and large-scale deployment.

To address this challenge without compromising output quality, {\bf speculative decoding} \cite{chenAcceleratingLargeLanguage2023,leviathanFastInferenceTransformers2023} has emerged as a leading lossless acceleration framework. The key idea is to employ a lightweight and fast \emph{draft model} to generate a sequence of candidate tokens, which are then verified in parallel by the powerful \emph{target LLM}. Extensions such as {\bf Medusa}~\cite{caiMedusaSimpleLLM2024a,ankner2024hydra,liu2024parallel} predict multiple candidate tokens from the target model's hidden state using draft heads, while more advanced methods like {\bf EAGLE-1/2/3}~\cite{liEAGLESpeculativeSampling2024,liEAGLE2FasterInference2024,liEAGLE3ScalingInference2025} operate at the feature level rather than the token level, developing single-layer draft models that achieve notable speedups.

Despite these advances, a critical bottleneck remains: the drafting process itself is still autoregressive. To generate a draft sequence of $N$ tokens, even highly efficient methods like EAGLE must perform $N$ sequential forward passes. While each pass is lightweight, their cumulative latency limits further acceleration, especially for longer drafts.

\begin{figure}[htb]
      \centering
      \centerline{\includegraphics[width=8.5cm]{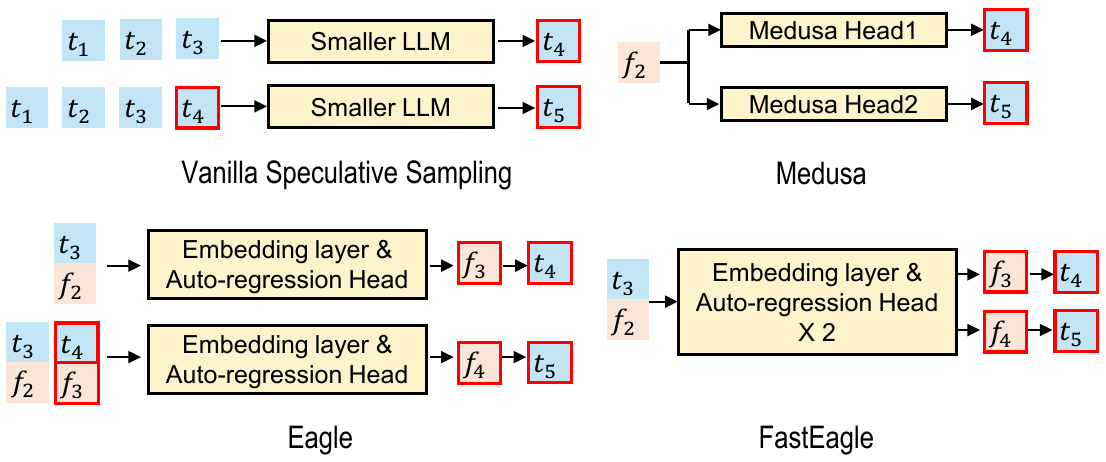}}
      \caption{Comparison of drafting strategies for the 4th and 5th tokens ($t_4, t_5$). Blue blocks ($t$) denote tokens, orange blocks ($f$) represent features with subscripts marking their sequence positions, and red borders highlight draft model predictions.}
      \label{fig:drafting}
\end{figure}

In this paper, we directly confront this limitation by asking: {\bf can we break the autoregressive chain within the drafting stage itself ?} We introduce {\bf FastEagle}, a novel non-autoregressive cascaded draft model that generates an entire draft sequence in a {\bf single forward pass}. Instead of temporal autoregression, our architecture employs a structural cascade of lightweight layers, where each layer predicts the hidden state of the next token in the sequence. To mitigate compounding errors inherent in such feed-forward structures, we design a {\bf multi-level supervision strategy} to ensure stable and accurate training. Fig.~\ref{fig:drafting} highlights the difference from prior methods. Our contributions are summarized as follows:
\begin{itemize}
    \item We present a cascaded non-autoregressive drafter that generates $N$ token distributions in one go, removing the need for $N$ sequential steps.
    \item We design a multi-level training approach with layer-specific supervision and feature alignment to ensure stable predictions and reduce error spread.
    \item We implement a simple constrained draft tree for linear verification complexity. Tests on various LLMs and tasks demonstrate consistent speedups over EAGLE-3.
\end{itemize}

\section{Methodology}
\subsection{FastEagle}
\label{sec:FastEagle_architecture}
Our approach builds upon the speculative decoding paradigm, a lossless \emph{draft-then-verify} framework for accelerating LLM inference. SOTA drafters such as \textbf{EAGLE} achieve high efficiency by employing a single lightweight decoder layer that operates directly on the target model's hidden features. However, this design remains \emph{autoregressive}: generating a draft of length $N$ requires $N$ sequential forward passes, where the feature $\mathbf{f}_{t+i}$ is conditioned on $\mathbf{f}_{t+i-1}$. This sequential dependency introduces a latency bottleneck that limits scalability.

\begin{figure}[htb]
      \centering
      \centerline{\includegraphics[width=8.5cm]{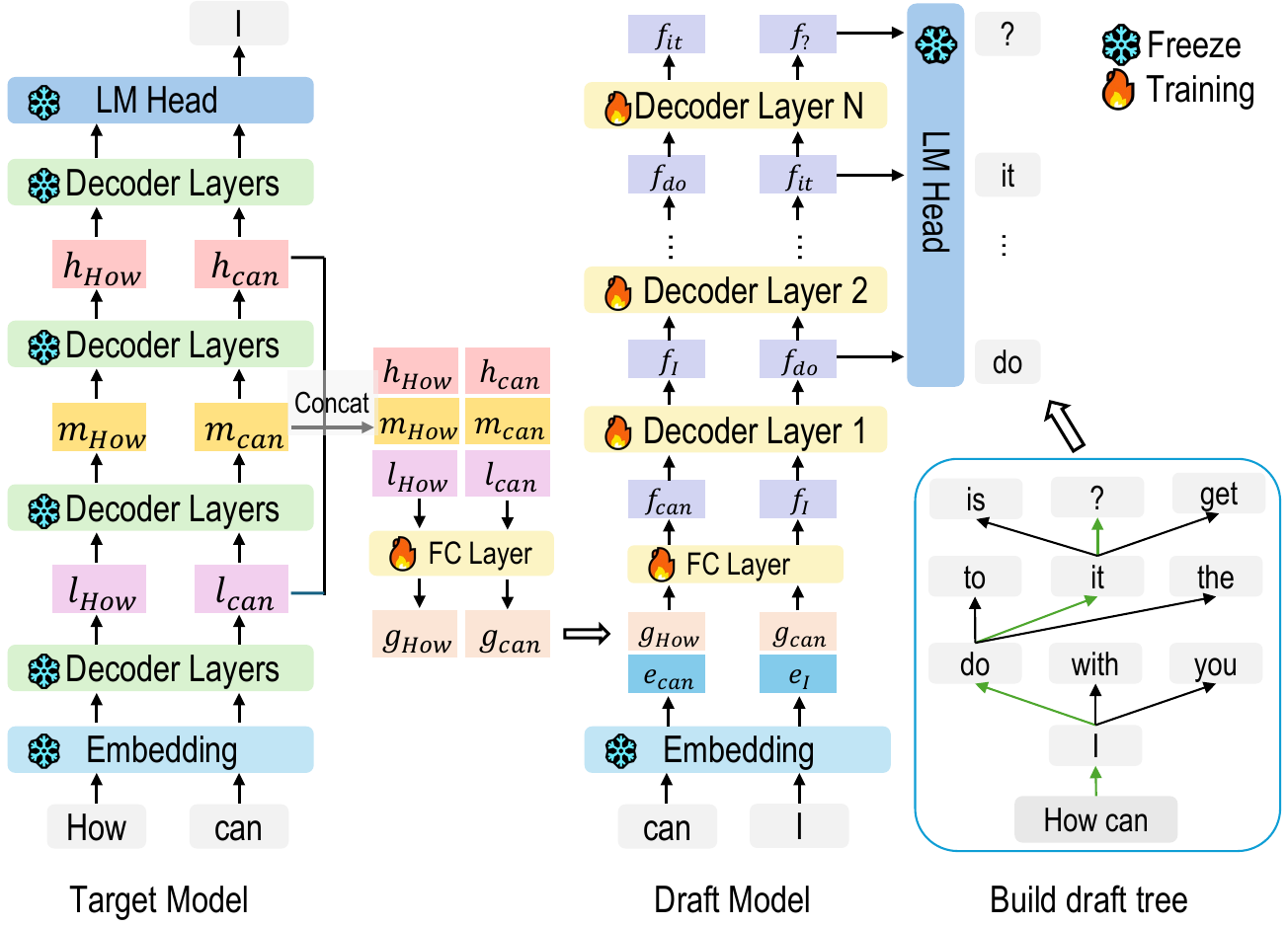}}
      \caption{Illustration of the FastEagle architecture: a cascaded parallel drafting framework that replaces autoregressive steps with layer-wise predictions, enabling single-pass generation of multiple future tokens while preserving output fidelity.}
      \label{fig:FastEagle_frame}
\end{figure}

To remove this bottleneck, we propose the \textbf{Cascaded Drafter}, a novel feed-forward architecture that replaces temporal autoregression with a deep structural cascade. As illustrated in Fig.~\ref{fig:FastEagle_frame}, the drafter comprises $N$ lightweight decoder layers, $L_1, \dots, L_N$, connected in series. Following the design philosophy of EAGLE-3~\cite{liEAGLE3ScalingInference2025}, the input is derived from the last verified tokens of the previous step. Specifically, we extract multi-level features $\mathbf{l}_t, \mathbf{m}_t, \mathbf{h}_t$ from the target model, concatenate them, and transform them via a fully connected (FC) layer to produce a fused feature $\mathbf{g}_t$. This feature is concatenated with the embedding $\mathbf{e}_{t+1}$ of the next token, forming the initial input to the cascade:
{\small
\begin{align}
      \mathbf{h}_1 = L_1(\text{FC}(\text{concat}(\mathbf{g}_t, \mathbf{e}_{t+1}))) \\
      \mathbf{h}_i = L_i(\mathbf{h}_{i-1}), \quad i = 2, \dots, N
\end{align}
}Each decoder layer $L_i$ maintains its own parameters, enabling specialization across cascade depth. The hidden state $\mathbf{h}_i$ at layer $i$ is projected through the frozen LM Head of the target model to produce the token distribution for position $t+i$. This \emph{layer-wise supervision} allows shallower layers to focus on short-range predictions while deeper layers capture longer-range dependencies. Unlike parallel-head methods, our cascaded design leverages intermediate computations hierarchically, enabling richer modeling of sequential structure while completing the full draft in a single forward pass.

\subsection{Constrained Draft Tree Construction}
\label{sec:constrained_draft_tree_construction}
Naively expanding the $N$ output distributions into a Cartesian product leads to exponential growth ($k^N$ paths for top-$k$ sampling), making verification infeasible. We therefore adopt a simple yet effective \textbf{Backbone Expansion} that constrains complexity to linear scale. We first sample $k$ candidates from $\mathbf{q}_{t+1}$; the most probable token becomes the backbone node and the remaining $k-1$ tokens become side branches. For each subsequent step $i=2,\dots,N$, we again sample $k$ tokens from $\mathbf{q}_{t+i}$ and attach them as children of the previous backbone node, extending the backbone with the most probable token. This produces exactly one backbone path of length $N$ and at most $k-1$ side branches per level, yielding $\mathcal{O}(Nk)$ nodes (approximately $1 + (N-1)\times k$). In practice this keeps tree-attention verification and memory cost linear in both depth and top-$k$, while preserving diversity near the backbone; when $k{=}1$ it deterministically degenerates to a chain.

\subsection{Draft Model Training}
Training the proposed cascaded drafter introduces a key challenge: errors from shallow layers may propagate through the cascade, destabilizing training and degrading long-range predictions. To address this, we design a \textbf{multi-level training objective} that supervises each layer directly. This ensures that every layer is optimized both for its own prediction and for producing stable representations for deeper layers. The overall loss $L_{\text{total}}$ is a weighted sum of per-layer losses:
{\small
\begin{equation}
L_{\text{total}} = \sum_{i=1}^{N} w_i \cdot (\alpha L_{\text{CE}, i} + \beta L_{\text{feat}, i}),
\end{equation}
}where $w_i$ are layer weights, and $\alpha, \beta$ balance token prediction and feature alignment.

\textbf{Token Prediction Loss ($L_{\text{CE}, i}$):} The primary objective is cross-entropy between the drafter's predicted distribution $\mathbf{q}_{t+i}$ and the target model's teacher distribution $\mathbf{p}_{t+i}$:
{\small
\begin{equation}
L_{\text{CE}, i} = - \sum_{k=1}^{V} (\mathbf{p}_{t+i})_k \log (\mathbf{q}_{t+i})_k,
\end{equation}
}where $V$ is the vocabulary size.

\textbf{Feature Alignment Loss ($L_{\text{feat}, i}$):} As a regularizer, this term minimizes the distance between the hidden state $\mathbf{h}_i \in \mathbb{R}^d$ of drafter layer $L_i$ and the corresponding feature $\mathbf{f}_i$ from the target LLM. Following \cite{liEAGLE3ScalingInference2025}, we adopt Smooth L1:
{\small
\begin{equation}
L_{\text{feat}, i} = \sum_{j=1}^{d} \text{SmoothL1}((\mathbf{h}_i)_j - (\mathbf{f}_i)_j),
\end{equation}
}with
{\small
\begin{equation}
\text{SmoothL1}(x) =
\begin{cases}
0.5 x^2 & |x| < 1, \\
|x| - 0.5 & \text{otherwise}.
\end{cases}
\end{equation}
}This loss anchors the drafter's hidden states to the target feature manifold, ensuring stable cascaded inputs. Crucially, training is \textbf{end-to-end} without teacher forcing: each $L_i$ consumes $\mathbf{h}_{i-1}$ from the same forward pass, exposing the model to inference-like errors during training. In our experiments, we use $w_i = 0.9^{N-i}$, $\alpha = 0.1$, and $\beta = 1.0$ in line with~\cite{liEAGLE3ScalingInference2025}.

\subsection{Inference Pipeline}
\label{sec:inference_pipeline}
We integrate the Cascaded Drafter into the speculative decoding loop. Each generation cycle proceeds as follows:

\textbf{Non-Autoregressive Drafting.}  
Given the last verified token $t$, the drafter produces $N$ distributions $\{\mathbf{q}_{t+i}\}_{i=1}^N$ in a single pass. A constrained draft tree is then built via the Backbone Expansion strategy (Sec.~\ref{sec:constrained_draft_tree_construction}).

\textbf{Parallel Verification.}  
The draft tree is passed to the target model $M_{\text{target}}$. Using tree attention, the model verifies all nodes in parallel, and the speculative sampling algorithm identifies the longest accepted token path.

\textbf{Update and Iterate.}  
Accepted tokens are appended to the output, and the target LLM's KV cache is updated. The features of the last accepted token initialize the next drafting cycle, and the process repeats.

\begin{table*}[htb]
      \centering
      \caption{Comparison of speedup ratios and average acceptance lengths $\tau$ across various methods. V stands for Vicuna, L31 for LLaMA-Instruct 3.1, L33 for LLaMA-Instruct 3.3, and DSL for DeepSeek-R1-Distill-LLaMA. Methods such as Medusa, which relax acceptance criteria in non-greedy scenarios, do not ensure lossless speedup. Hence, similar to EAGLE-3, FastEagle is not compared with these methods at Temperature=1.}
      \resizebox{0.85\linewidth}{!}{
        \begin{tabular}{cccccccccccccc}
        \toprule
              &       & \multicolumn{2}{c}{MT-Bench} & \multicolumn{2}{c}{HumanEval} & \multicolumn{2}{c}{GSM8K} & \multicolumn{2}{c}{Alpaca} & \multicolumn{2}{c}{CNN/DM} & \multicolumn{2}{c}{Mean} \\
        \midrule
        Model & Method & Speedup & $\tau$     & Speedup & $\tau$     & Speedup & $\tau$     & Speedup & $\tau$     & Speedup & $\tau$     & Speedup & $\tau$ \\
        \midrule
        \multicolumn{14}{c}{Temperature=0} \\
        \midrule
        \multirow{4}[2]{*}{V 13B} 
              & SpS       & 1.93x & 2.27  & 2.23x & 2.57  & 1.77x & 2.01  & 1.76x & 2.03  & 1.93x & 2.33  & 1.92x & 2.24 \\
              & Medusa    & 2.07x & 2.59  & 2.50x & 2.78  & 2.23x & 2.64  & 2.08x & 2.45  & 1.71x & 2.09  & 2.12x & 2.51 \\
              & EAGLE-3   & 5.08x & \textbf{6.25}  & 5.97x & \textbf{7.14}  & 4.82x & \textbf{5.89}  & 4.66x & \textbf{5.77}  & 4.51x & \textbf{6.07}  & 5.01x & \textbf{6.22} \\
              & FastEagle & \textbf{5.76x} & 6.20 & \textbf{6.64x} & 7.10 & \textbf{5.51x} & 5.85 & \textbf{5.32x} & 5.73 & \textbf{5.20x} & 6.03 & \textbf{5.67x} & 6.18 \\
        \midrule
        \multirow{2}[2]{*}{L31 8B} 
              & EAGLE-3   & 3.90x & \textbf{5.83}  & 4.35x & \textbf{6.34}  & 3.98x & \textbf{5.83}  & 4.32x & \textbf{6.30}  & 3.15x & \textbf{4.94}  & 3.94x & \textbf{5.85} \\
              & FastEagle & \textbf{4.73x} & 5.68 & \textbf{5.16x} & 6.30 & \textbf{4.76x} & 5.79 & \textbf{5.13x} & 6.26 & \textbf{3.91x} & 4.90 & \textbf{4.74x} & 5.79 \\
        \midrule
        \multirow{2}[1]{*}{L33 70B} 
              & EAGLE-3   & 3.61x & \textbf{5.33}  & 4.29x & \textbf{6.13}  & 3.89x & \textbf{5.75}  & 3.85x & \textbf{5.69}  & 2.77x & \textbf{4.75}  & 3.62x & \textbf{5.53} \\
              & FastEagle & \textbf{4.44x} & 5.18 & \textbf{5.07x} & 6.08 & \textbf{4.64x} & 5.71 & \textbf{4.60x} & 5.65 & \textbf{3.57x} & 4.68 & \textbf{4.46x} & 5.46 \\
        \midrule
        \multirow{2}[1]{*}{DSL 8B} 
              & EAGLE-3   & 3.55x & \textbf{5.28}  & 4.09x & \textbf{5.98}  & 4.51x & \textbf{6.63}  & 3.15x & \textbf{5.07}  & 3.02x & \textbf{4.62}  & 3.66x & \textbf{5.52} \\
              & FastEagle & \textbf{4.32x} & 5.23 & \textbf{4.87x} & 5.94 & \textbf{5.32x} & 6.59 & \textbf{3.96x} & 5.03 & \textbf{3.83x} & 4.58 & \textbf{4.46x} & 5.47 \\
        \midrule
        \multicolumn{14}{c}{Temperature=1} \\
        \midrule
        \multirow{3}[2]{*}{V 13B} 
              & SpS       & 1.62x & 1.84  & 1.72x & 1.97  & 1.46x & 1.73  & 1.52x & 1.78  & 1.66x & 1.89  & 1.60x & 1.84 \\
              & EAGLE-3   & 4.07x & \textbf{5.12}  & 4.65x & \textbf{5.92}  & 4.21x & \textbf{5.28}  & 3.99x & \textbf{5.09}  & 3.83x & \textbf{5.42}  & 4.15x & \textbf{5.37} \\
              & FastEagle & \textbf{4.85x} & 5.07 & \textbf{5.43x} & 5.88 & \textbf{5.03x} & 5.24 & \textbf{4.78x} & 5.05 & \textbf{4.62x} & 5.38 & \textbf{4.94x} & 5.32 \\
        \midrule
        \multirow{2}[2]{*}{L31 8B} 
              & EAGLE-3   & 2.57x & \textbf{3.94}  & 3.63x & \textbf{5.42}  & 2.82x & \textbf{4.29}  & 3.40x & \textbf{5.16}  & 2.49x & \textbf{4.09}  & 2.95x & \textbf{4.58} \\
              & FastEagle & \textbf{3.35x} & 3.89 & \textbf{4.41x} & 5.38 & \textbf{3.65x} & 4.25 & \textbf{4.27x} & 5.12 & \textbf{3.32x} & 4.05 & \textbf{3.80x} & 4.54 \\
        \midrule
        \multirow{2}[2]{*}{L33 70B} 
              & EAGLE-3   & 3.46x & \textbf{5.15}  & 3.86x & \textbf{5.86}  & 3.67x & \textbf{5.65}  & 3.64x & \textbf{5.57}  & 2.61x & \textbf{4.68}  & 3.45x & \textbf{5.38} \\
              & FastEagle & \textbf{4.25x} & 5.10 & \textbf{4.98x} & 5.82 & \textbf{4.74x} & 5.61 & \textbf{4.72x} & 5.53 & \textbf{3.73x} & 4.64 & \textbf{4.48x} & 5.34 \\
        \midrule
        \multirow{2}[2]{*}{DSL 8B} 
              & EAGLE-3   & 2.70x & \textbf{4.19}  & 3.27x & \textbf{4.88}  & 3.88x & \textbf{5.70}  & 2.66x & \textbf{4.00}  & 2.58x & \textbf{3.97}  & 3.02x & \textbf{4.52} \\
              & FastEagle & \textbf{3.51x} & 4.14 & \textbf{4.08x} & 4.84 & \textbf{4.67x} & 5.66 & \textbf{3.45x} & 3.96 & \textbf{3.36x} & 3.93 & \textbf{3.81x} & 4.51 \\
        \bottomrule
        \end{tabular}%
        }
      \label{tab:experiments}%
\end{table*}%

\section{Experiments}
\label{sec:experiments}
We conduct extensive experiments to evaluate the effectiveness of FastEagle. 

\textbf{Setup.} We use LLaMA-Instruct 3.1 8B \cite{dubey2024llama} and Vicuna-13B \cite{chiang2023vicuna} as target models. Tasks include multi-turn dialogue (MT-Bench \cite{zheng2023judging}), code generation (HumanEval \cite{chen2021evaluating}), and mathematical reasoning (GSM8K \cite{cobbe2021training}). Draft models are trained on ShareGPT \cite{ding2023enhancing} in a zero-shot setting, consistent with prior lightweight drafters such as EAGLE and Medusa.

\textbf{Baselines.} We compare against Vanilla Autoregressive decoding, Standard Speculative Sampling (SpS) \cite{chen2021evaluating,leviathanFastInferenceTransformers2023}, Medusa \cite{caiMedusaSimpleLLM2024a}, and EAGLE-3 \cite{liEAGLE3ScalingInference2025}. Since FastEagle is a \emph{lossless} acceleration method, we do not assess generation quality.

\textbf{Metrics.} We report (i) \emph{Speedup Ratio}, the wall-clock speedup over vanilla decoding, and (ii) \emph{Average Acceptance Length} $\tau$, the average number of tokens accepted per verification step. Results are reported under both greedy decoding (Temperature=0) and stochastic sampling (Temperature=1).

\textbf{Implementation.} Similar to EAGLE-3, we use the AdamW optimizer with $(\beta_1,\beta_2)=(0.9,0.95)$ and gradient clipping of 0.5. The learning rate is 5e-5. We use ShareGPT and UltraChat-200K~\cite{ding2023enhancing} as training data (approximately 68K and 464K entries). We call the target model to generate responses rather than using a fixed dataset. For the reasoning model DeepSeek-R1-Distill-LLaMA 8B, we also use the OpenThoughts-114k-math dataset. Training runs on 8\,\texttimes\,NVIDIA A100-80G GPUs. We evaluate on NVIDIA A100-80G. All latency is measured on a single GPU for 7B--13B and two GPUs for 70B with batch size $=1$ for consistency. Following EAGLE-3, the draft uses tree attention with Top-$K=10$, depth $=7$.  

\subsection{Main Results}
\label{sec:main_results}

Table~\ref{tab:experiments} summarizes the results. Across all target models and datasets, FastEagle consistently outperforms strong autoregressive drafters in wall-clock speedup while maintaining comparable average acceptance lengths $\tau$. 
At Temperature=0, mean speedups improve from 5.01$\times$\,$\to$\,5.67$\times$ (Vicuna-13B), 3.94$\times$\,$\to$\,4.74$\times$ (LLaMA-Instruct 3.1 8B), 3.62$\times$\,$\to$\,4.46$\times$ (LLaMA-Instruct 3.3 70B), and 3.66$\times$\,$\to$\,4.46$\times$ (DSL 8B), with $\tau$ within $\pm$0.1 of EAGLE-3 (6.22 vs 6.18; 5.85 vs 5.79; 5.53 vs 5.46; 5.52 vs 5.47). Under Temperature=1, the trend holds: 4.15$\times$\,$\to$\,4.94$\times$ (Vicuna-13B), 2.95$\times$\,$\to$\,3.80$\times$ (LLaMA-Instruct 3.1 8B), 3.45$\times$\,$\to$\,4.48$\times$ (LLaMA-Instruct 3.3 70B), and 3.02$\times$\,$\to$\,3.81$\times$ (DSL 8B), again with $\tau$ within $\pm$0.06. 
These results indicate that eliminating the $N$-step drafting latency via a single-pass cascade yields consistent end-to-end gains while preserving lossless verification behavior.

\begin{table}[htb]
  \caption{Ablation study results with LLaMA-Instruct 8B as the target model at Temperature=0. ``w/o Constrained Tree" replaces the draft tree with a simple chain. "w/o Cascaded Structure" uses a parallel instead of a serial architecture. "w/o Feature Loss" is trained only with cross-entropy loss.}
  \label{tab:ablation}
  \centering
  \resizebox{0.85\linewidth}{!}{
  \begin{tabular}{l|cc|cc}
    \hline
    \multirow{2}{*}{Method} & \multicolumn{2}{c|}{MT-Bench} & \multicolumn{2}{c}{GSM8K} \\
    \cline{2-5}
    & Speedup & $\tau$ & Speedup & $\tau$ \\
    \hline
    \textbf{Our Method (Full)} & \textbf{4.40x} & \textbf{6.13} & \textbf{4.48x} & \textbf{6.23} \\
    w/o Constrained Tree & 3.81x & 5.15 & 3.85x & 5.21 \\
    w/o Cascaded Structure & 3.52x & 4.81 & 3.60x & 4.95 \\
    w/o Feature Loss & 3.19x & 4.07 & 3.25x & 4.18 \\
    \hline
  \end{tabular}
  }
  \vspace{-0.3cm}
\end{table}

\subsection{Ablation Studies}
\label{sec:ablation_studies}
\textbf{Different components of FastEagle.}
We quantify each component's contribution on LLaMA-Instruct 8B at Temperature=0 (Table~\ref{tab:ablation}). Removing the constrained tree (``w/o Constrained Tree'') reduces speedup from 4.40$\times$ to 3.81$\times$ on MT-Bench ($-0.59\times$, $\approx$15\% relative) and from 4.48$\times$ to 3.85$\times$ on GSM8K ($-0.63\times$), while $\tau$ drops by $\approx$1.0 (6.13$\to$5.15; 6.23$\to$5.21). Replacing the cascaded structure with parallel heads (``w/o Cascaded Structure'') further degrades performance (3.52$\times$/4.81 vs. 4.40$\times$/6.13 on MT-Bench; 3.60$\times$/4.95 vs. 4.48$\times$/6.23 on GSM8K), underscoring the benefit of hierarchical refinement. The largest drop occurs without feature alignment loss (``w/o Feature Loss''), where speedup falls to 3.19$\times$/3.25$\times$ and $\tau$ declines by $\approx$2.0 (to 4.07/4.18), highlighting feature-level anchoring as essential for stabilizing cascaded predictions and mitigating error propagation.

\begin{figure}[htb]
      \centering
      \centerline{\includegraphics[width=6.5cm]{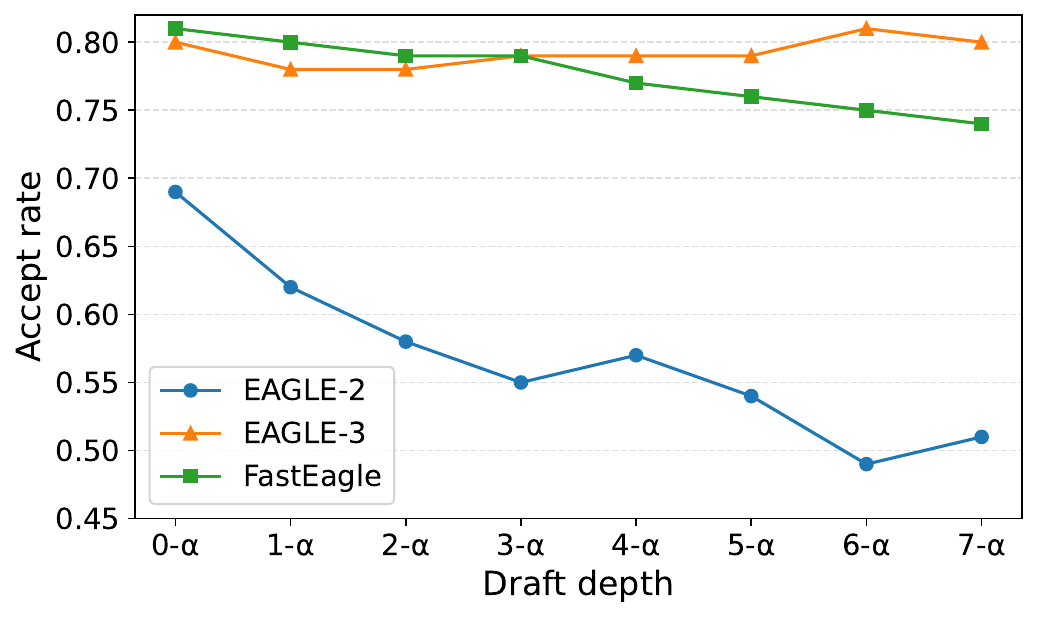}}
      \caption{Acceptance rates across methods on MT-Bench at T=0. FastEagle maintains a high and relatively stable acceptance with mild decline at deeper layers, EAGLE-3 remains the most stable overall, while EAGLE-2 exhibits  degradation.}
      \label{fig:acceptance_rate}
      \vspace{-0.3cm}
\end{figure}

\textbf{Comparison of accept rate.}
Fig.~\ref{fig:acceptance_rate} shows that FastEagle sustains high acceptance with a mild depth-wise decline (from 0.81 to 0.74 on MT-Bench), whereas EAGLE-3 is the most stable overall ($\approx$0.78--0.81) and EAGLE-2 degrades substantially (0.69 to 0.51). Together with Table~\ref{tab:ablation}, where removing the feature loss causes a $\tau$ drop of about 2.0, these results suggest that layer-wise supervision and feature alignment effectively curb long-range error accumulation in the cascaded setting.

\begin{table}[htb]
      \centering
      \caption{Throughput improvement in various batch sizes on A100 and LLaMA-Instruct 3.1 8B for MT-Bench with vLLM.}
      \resizebox{1.0\linewidth}{!}{
        \begin{tabular}{ccccccccc}
        \toprule
        Batch size & 2     & 4     & 8     & 16    & 24    & 32    & 48    & 56 \\
        \midrule
        EAGLE      & 1.30x & 1.25x & 1.21x & 1.10x & 1.03x & 0.93x & 0.82x & 0.71x \\
        EAGLE-3    & 1.75x & 1.68x & 1.58x & 1.49x & 1.42x & 1.36x & 1.21x & 1.01x \\
        FastEagle  & 1.95x & 1.88x & 1.78x & 1.69x & 1.62x & 1.22x & 0.97x & 0.88x \\
        \bottomrule
        \end{tabular}%
        }
      \label{tab:vllm_fasteagle}%
      \vspace{-0.5cm}
\end{table}%

\textbf{FastEagle in vLLM}
We further evaluate the impact of FastEagle on throughput under large batch sizes using vLLM~\cite{kwon2023efficient}, a widely used production-grade framework. Results on A100 with LLaMA-Instruct 3.1 8B are shown in Table \ref{tab:vllm_fasteagle}. EAGLE-3 achieves its peak improvement at batch size 56, while FastEagle peaks at 32. We attribute this to memory pressure at larger batch sizes: FastEagle maintains additional KV cache states, making it increasingly memory-bound as batch size grows. This study disables the tree structure, sets the maximum chain length to 2, and uses MT-Bench as the evaluation dataset. 

\section{Conclusion}
FastEagle is a non-autoregressive cascaded drafter that produces an entire draft in a single pass, removing the $N$-step drafting bottleneck. Across various models, it consistently outperforms EAGLE-3 in speedup under both greedy and stochastic decoding with comparable average acceptance length. Removing sequential dependencies in drafting offers a robust, scalable route to accelerate LLM inference.




\bibliographystyle{IEEEbib}
\bibliography{refs}

\end{document}